\documentclass[letterpaper]{article} 
\usepackage[preprint]{aaai2027}  
\usepackage[hyphens]{url}  
\usepackage{graphicx} 
\usepackage{natbib}  
\usepackage{caption} 
\usepackage{algorithm}
\usepackage{algorithmic}

\usepackage{newfloat}
\usepackage{listings}
\DeclareCaptionStyle{ruled}{labelfont=normalfont,labelsep=colon,strut=off} 
\floatstyle{ruled}
\newfloat{listing}{tb}{lst}{}
\floatname{listing}{Listing}

\usepackage{booktabs}

\usepackage{amsmath}
\usepackage{amssymb}
\usepackage{comment}
\usepackage{multirow}

\title{LUCID: Latent-Skill Unified Control via Imagined Dynamics\\
for Long-Horizon Humanoid Loco-Manipulation}

\author{
    Cheng Guo\textsuperscript{\rm 1,\rm 2}\corresponding,
    Mingzhe Ni\textsuperscript{\rm 1},
    Angelo Cangelosi\textsuperscript{\rm 1},
    Arash Ajoudani\textsuperscript{\rm 2}
}

\affiliations{
    \textsuperscript{\rm 1}Department of Computer Science, University of Manchester, Manchester, UK\\

    \textsuperscript{\rm 2}Human-Robot Interfaces and Interaction Laboratory, Italian Institute of Technology, Genoa, Italy\\
}

\begin{document}

\maketitle

\begin{abstract}
Long-horizon humanoid loco-manipulation requires composing versatile whole-body skills and reliable high-level decision making. Existing methods often coordinate pretrained skills with scripted planners, finite-state machines or task-specific model-free policies, restricting their ability to handle complex task sequences. 
To address this limitation, we propose \textbf{LUCID}, a hierarchical model-based reinforcement learning framework that plans over reusable skills through imagined rollouts of a learned dynamics model. 
LUCID first trains a structured latent-conditioned low-level policy via adversarial imitation and then freezes it while jointly learning a high-level policy and macro-dynamics world model. 
The world model predicts the temporally extended state transitions induced by latent decisions, enabling high-level policy optimization through imagined rollouts.
We evaluate our framework across various simulated multi-object rearrangement scenarios. Experimental results show that LUCID improves the full-task success and partial-completion rates compared to prior baseline methods, demonstrating its effectiveness in complex sequential loco-manipulation tasks.
\end{abstract}

\section{Introduction}
Humanoids can perform individual locomotion and manipulation behaviors, but executing them reliably in a long-horizon sequence remains a challenge~\cite{gu2026humanoid}. In multi-object rearrangement, for example, a humanoid needs to repeatedly navigate between objects and manipulate them within a continuous episode. Such tasks require both versatile whole-body skills and a high-level policy that can coordinate them toward sequential goals. 
Previous works have made substantial progress in physics-based humanoid control through adversarial imitation and latent skill learning, which enable natural whole-body motion and reusable motor behaviors~\cite{peng2021amp, peng2022ase, weng2025hdmi, wang2025physhsi}. Recent humanoid-scene interaction methods further extend such motor control to navigation and object manipulation in more complex environments~\cite{starke2019neural, hassan2023synthesizing, pan2024synthesizing, xiao2024unified}. Together, these works have expanded humanoid control from isolated motor skills to scene-aware loco-manipulation.

However, coordinating these behaviors over a long task sequence remains difficult. Existing approaches either consider primarily specialized controllers for isolated object interactions~\cite{xu2024humanvla, wang2025skillmimic}, or connect consecutive interactions through hand-designed mechanisms, such as a fixed navigation-to-manipulation state machine~\cite{pan2025tokenhsi} or per-object reference tracking with scripted handoffs between stages~\cite{xu2025intermimic}. 
Although such systems handle individual interactions effectively, their temporal organization is largely predefined and does not account for how intermediate decisions affect later subtasks.
Hierarchical reinforcement learning addresses this structural problem by separating fast motor control from slower decision-making over temporally extended skills~\cite{sutton1999between}. In physics-based character control, latent-skill methods have produced controllable motion representations and compact categorical priors that can be reused for downstream tasks~\cite{tessler2023calm, zhu2023neural}. High-level policies can also combine pretrained humanoid motor primitives with limited task-specific reward design~\cite{kuang2025skillblender}. 
Yet existing hierarchical systems still rely on predefined skill transitions or task-specific model-free high-level policies trained through direct environment interaction. Consequently, they can select reusable behaviors but cannot predict how one skill changes the conditions for subsequent interactions, which is a limitation that becomes increasingly important in long-horizon multi-object rearrangement.

World models offer this predictive capability by approximating environment dynamics and using imagined trajectories to improve decision-making~\cite{ha2018world, hafner2020mastering}. They have demonstrated strong performance across control domains and have recently been applied to robotic manipulation, navigation, and locomotion~\cite{wu2023daydreamer, li2025robotic}. For humanoid interaction, however, predicting every joint-level transition over an extended horizon is difficult: high-dimensional, contact-rich dynamics require many autoregressive steps, amplifying model errors. The hierarchy provides a natural temporal abstraction. Rather than reproducing complete physical trajectories, a model can predict the task-level changes induced by each temporally extended skill. This connects reusable motor behaviors with predictive high-level reasoning, motivating a framework that learns skill-level dynamics and uses them to coordinate sequential humanoid interactions.

\begin{figure*}[t]
\centering
\includegraphics[width=0.95\textwidth]{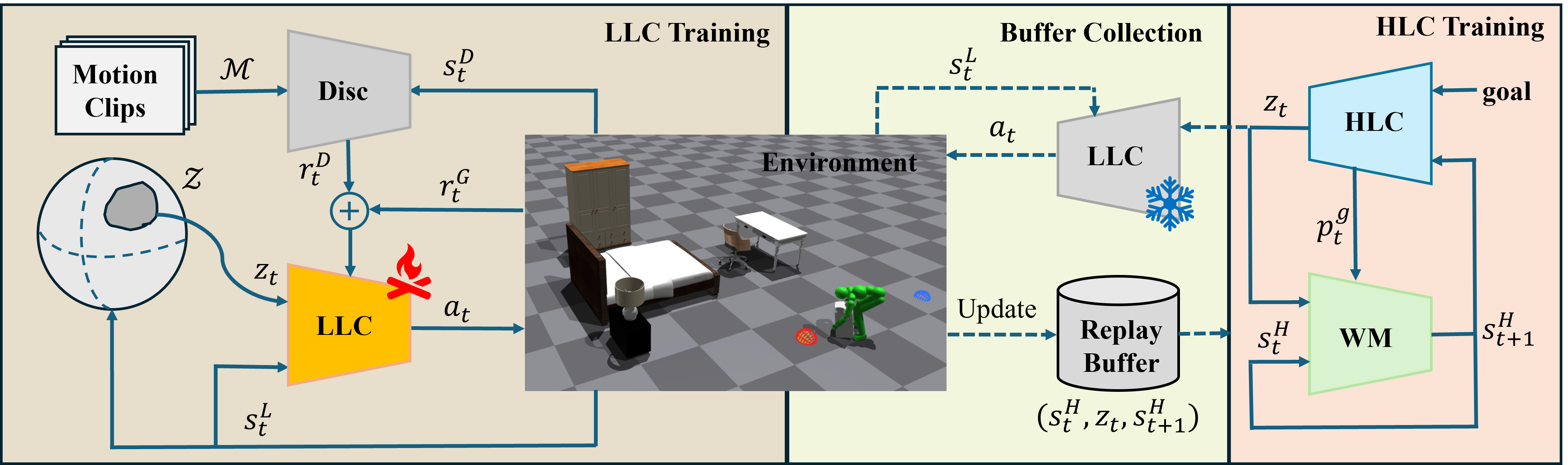} 
\caption{Overview of LUCID. LLC is trained with adversarial imitation and task rewards, then frozen. During training, latent z is produced with a state-based oracle. HLC is jointly trained with the macro-dynamics world model. The replay buffer is initialized with real transitions and refreshed during each update. Dashed lines indicate detached (stop-gradient) data flow.}
\label{fig:lucid_framework}
\end{figure*}

To this end, we introduce \textbf{LUCID}, a hierarchical model-based reinforcement learning framework for long-horizon skill composition, illustrated in Figure~\ref{fig:lucid_framework}. Unlike hierarchical humanoid controllers that rely on scripted transitions or task-specific model-free skill sequencing, LUCID learns skill-level dynamics and optimizes a goal-conditioned high-level policy through imagined macro-transitions. A frozen latent-conditioned low-level controller provides reusable whole-body skills through a structured interface combining discrete skill anchors with continuous variation. At the macro timescale, the high-level policy selects latent commands, while a macro-dynamics world model predicts their task-level consequences for the humanoid, manipulated objects, and task progress rather than modeling every joint-level transition. Imagined rollouts through this model allow the policy to anticipate how current skill choices affect subsequent subtasks, while real simulator experience continually improves the model. This combination enables LUCID to coordinate complete multi-object rearrangement sequences without scripted handoffs.

In summary, our contributions are threefold. 
\begin{itemize}
    \item We propose \textbf{LUCID}, a hierarchical model-based reinforcement learning framework for task planning from imagined dynamics. Its world model predicts how latent decisions shape task progression, allowing a high-level policy to optimize long-horizon behavior in imagination.
    \item We train a low-level controller with adversarial imitation and an interaction-aware curriculum, using a structured latent interface that exposes whole-body motor primitives to the high-level policy.
    \item We evaluate our method in various long-horizon multi-object rearrangement tasks, showing improved full sequence completion and validating the key component through ablations.
\end{itemize}

\section{Related Work}

\subsection{Physics-Based Motion Generation}
Physics-based motion generation produces dynamically feasible behaviors by training controllers in simulation. DeepMimic popularized reference-tracking reinforcement learning for character animation~\cite{peng2018deepmimic}, while AMP acquired task-compatible motion priors from unstructured demonstrations~\cite{peng2021amp}. Building on these foundations, reusable and directable representations enabled broader downstream applications~\cite{peng2022ase, tessler2023calm, tessler2024maskedmimic}. Interaction-oriented research further expanded the scope to physical scene contacts, dynamic imitation, and object grasping~\cite{hassan2023synthesizing, wang2023physhoi, luo2024omnigrasp}; general rearrangement~\cite{xu2024humanvla}; and scalable multi-skill execution~\cite{xu2025intermimic, pan2025tokenhsi}. Recent systems have also transferred such capabilities to real humanoids~\cite{wang2025physhsi, li2026haic}. 
Unlike these methods, LUCID organizes phase-controllable skills with a high-level controller trained in imagination.

\subsection{Hierarchical Reinforcement Learning}
Hierarchical reinforcement learning decomposes long-horizon control into high-level decisions and low-level skills~\cite{sutton1999between, bacon2017option}. Early systems built skill libraries and used schedulers to select or compose controllers~\cite{faloutsos2001composable, heess2016learning}. Later methods learned latent skill spaces that high-level policies could steer to generate motions~\cite{merel2018neural, peng2019mcp, hasenclever2020comic, tessler2023calm}. These abstractions were extended to legged loco-manipulation through skill sequencing and residual composition~\cite{ji2022hierarchical, kumar2023cascaded}. Recent humanoid systems instead plan motion references with hierarchical world models or blend goal-conditioned skills~\cite{hansen2025hierarchical, kuang2025skillblender}. LUCID learns discrete phase decisions over a frozen structured interface and optimizes long-horizon skill sequencing through imagined macro-dynamics for multi-object rearrangement.

\subsection{World Model for Robotics}
World models learn action-conditioned dynamics for planning and policy optimization through imagined trajectories~\cite{ha2018world}. PlaNet introduced latent dynamics for planning~\cite{hafner2019dream}, while Dreamer trained actor–critic policies from latent imagination~\cite{hafner2020mastering, hafner2025mastering}. TD-MPC combined learned representations with trajectory optimization~\cite{hansen2022temporal, hansen2024td}, and DayDreamer extended imagination-based learning to physical robots~\cite{wu2023daydreamer}. More recent robotics-specific models emphasize predictive reliability. RWM uses dual-autoregressive training to reduce compounding error and optimize locomotion policies in a learned neural simulator~\cite{li2025robotic}. HAIC instead predicts unobserved object dynamics from proprioception to improve feedback control during humanoid–object interaction~\cite{li2026haic}. However, these approaches primarily model joint-level dynamics, provide state estimates, or address continuous control. LUCID instead models skill-level transitions induced by structured phase commands and trains a high-level policy through imagined rollouts, enabling long-horizon planning while keeping the physics-based low-level controller fixed.

\section{Method}
\label{sec:method}
Figure~\ref{fig:lucid_framework} presents LUCID, a scalable hierarchical framework for long-horizon robot-object interaction tasks. Training proceeds in two stages. First, we train a reusable latent-conditioned low-level controller (LLC) via adversarial imitation following ASE~\cite{peng2022ase}. We then freeze the LLC and jointly train a macro-dynamics world model (WM) and a goal-conditioned high-level controller (HLC) similar to DreamerV3~\cite{hafner2025mastering}. The WM predicts the physical consequences of latent decisions, allowing the HLC to learn from imagined macro-rollouts while real simulator interaction continually supplies training data for the WM.
\subsection{Problem Formulation}
\label{sec:task_setup}
We consider a simulated humanoid that must rearrange multiple objects in a prescribed order. We formulate the task as a goal-conditioned Markov Decision Process (MDP)
$\langle\mathcal{S},\mathcal{A},\mathcal{T},\mathcal{R},\gamma,\mathcal{G}\rangle$.
At simulation step $t$, $s_t\in\mathcal{S}$ contains the humanoid and environment states, $a_t\in\mathcal{A}$ denotes joint-level PD targets, and $\mathcal{T}$ is induced by the physics simulator. The episode goal $g=(g^1,\ldots,g^M)\in\mathcal{G}$ specifies an ordered sequence of $M$ object-rearrangement subgoals. The objective is to maximize $\mathbb{E}[\sum_{t=0}^{T-1}\gamma^t r_t]$, where $r_t$ rewards subgoal completion and physically plausible motion.

\subsection{Latent Skill Primitives}
\label{sec:llc}
The LLC $\pi_L(a_t\mid s_t^L,z_t)$ maps a low-level state $s_t^L$ and a skill latent $z_t$ to joint-level actions. The low-level state contains proprioception, object state, and the current task-guidance variables; in particular, the bounded guidance offset and waypoint-advance information represented by $p_t^g$ are encoded into $s_t^L$ before LLC execution. We train it to produce human-like, interaction-aware behaviors while remaining controllable through $z_t$ by combining an adversarial imitation reward $r_t^D$ with a goal-conditioned task reward $r_t^G$:
\begin{equation}
\max_{\pi_L}\;
\mathbb{E}_{\tau\sim p(\tau\mid\pi_L,\psi)}
\left[
\sum_{t=0}^{T-1}\gamma^t
\bigl(r_t^D+\lambda_G r_t^G\bigr)
\right].
\label{eq:llc_objective}
\end{equation}
Here, $\tau=(s_0^L,z_0,a_0,s_1^L,\ldots)$ is a low-level trajectory, $\lambda_G$ balances task progress and imitation, and the oracle $\psi(s_t^L,g)$ selects the skill associated with the current interaction phase. By consistently conditioning each phase on its corresponding latent during training, the LLC learns to produce distinct behaviors for different latent inputs, thereby providing latent-level behavior controllability.

\paragraph{Adversarial imitation}
Let $\phi_t=\phi(s_t^D,s_{t+1}^D)$ be a transition feature constructed from robot and object kinematics. A discriminator $D(\phi)\in(0,1)$ is trained to classify reference transitions as real and LLC transitions as generated. The LLC therefore receives the non-saturating reward:
\begin{equation}
 r_t^D=-\log\big(1-D(\phi_t)\big),
\label{eq:style_reward_main}
\end{equation}
which increases when a generated transition is assigned a higher reference-motion probability. The full discriminator objective, including its gradient regularizer, is provided in the supplementary material (\emph{Adversarial Imitation Objective}).

\paragraph{Structured latent space}
\label{sec:latent_space}
Unlike the unstructured hyperspherical prior used by ASE, we reserve one anchor for each of $N$ semantic skills and use the remaining dimensions for within-skill variation:
\begin{equation}
 z_t=
 \frac{[\,\mathbf{e}_{c_t};\;\sigma_z\boldsymbol{\epsilon}_t\,]}
 {\left\lVert[\,\mathbf{e}_{c_t};\;\sigma_z\boldsymbol{\epsilon}_t\,]\right\rVert_2},
 \quad
 \boldsymbol{\epsilon}_t\sim\mathcal{N}(\mathbf{0},\mathbf{I}_{D_z-N}),
\label{eq:structured_z}
\end{equation}
where $D_z\geq N$ is the latent dimension, $\mathbf{e}_{c_t}\in\mathbb{R}^{N}$ is the one-hot anchor for skill $c_t\in\{1,\ldots,N\}$, and $\sigma_z$ controls within-skill variation. During LLC training, $c_t$ is supplied by $\psi$, the induced latent distribution is thus a mixture $p(z_t)=\sum_{c=1}^{N}p(c_t=c)\,p(z_t\mid c_t=c)$ of $N$ such neighborhoods rather than a uniform distribution over the sphere. After the LLC is frozen, the HLC selects the structured latent.

\paragraph{Interaction-aware curriculum}
\label{sec:llc_training}
To maintain behavior controllability during sequential object rearrangement, we train $\pi_L$ using a three-stage curriculum: 1) \emph{carry}, which emphasizes approach, contact, and lifting; 2) \emph{rearrangement}, which enables transport and placement; and 3) \emph{retreat and chaining}, which adds the transit skill, a subtask-completion bonus, and a post-placement retreat term.
Each stage warm-starts from the previous one to target progressive completions, with only the task reward $r^G$ and reference-state initialization (RSI) changing. 
At every stage, the task reward is defined as:
\begin{equation}
 r_{t,m}^G=\sum_j w_j^{(m)} r_t^{(j)},
\label{eq:task_reward_main}
\end{equation}
where $m$ indexes the curriculum stage, $r_t^{(j)}$ is an interaction reward, and $w_j^{(m)}$ is its stage-dependent coefficient. The stage-specific objective, retreat reward, activation conditions, and initialization scheme are given in the supplementary material (\emph{Interaction-Aware Curriculum}).

\subsection{Macro-Dynamics World Model}
\label{sec:wm}
\paragraph{Macro transition}
The WM predicts task-level physical consequences over $K$ low-level control steps. At macro step $t$, the HLC chooses $a_t^H=(z_t,p_t^g)$, where $z_t$ is the structured skill latent and $p_t^g$ contains a bounded guidance offset and a waypoint-advance gate. Before LLC execution, $p_t^g$ is encoded into the task-guidance portion of $s_t^L$. The frozen LLC then executes the selected latent for up to $K$ simulator steps, producing one real macro transition.
We decompose the compact macro state as $s_t^H=(s_t^c,s_t^f)$, where $s_t^c$ contains continuous task variables (e.g., poses, velocities, displacements, lift heights) and $s_t^f$ contains binary task-progress flags. The dynamics network uses $(s_t^H,a_t^H)$ to predict the next continuous state and progress flags, whereas a separate continuation head estimates
whether the episode remains active:
\begin{equation}
\begin{aligned}
\hat{s}_{t+1}^{c}
 &=s_t^{c}+f_{\theta}(s_t^{H},a_t^{H}),\\
\hat{s}_{t+1}^{f}
 &=\operatorname{sigmoid}\!\big(
   g_{\theta}(s_t^{H},a_t^{H})\big),\\
\hat{\zeta}_{t}
 &=\operatorname{sigmoid}\!\big(h_{\theta}(s_t^{H})\big).
\end{aligned}
\label{eq:wm_pred}
\end{equation}
Here, $f_\theta$ predicts the change in the continuous variables rather than their absolute next values. This residual formulation focuses learning on the state change induced by one macro action. In contrast, $g_\theta$ directly predicts the probabilities of the binary flags because their abrupt changes, such as a placed flag switching from zero to one, are poorly represented by continuous residuals. The continuation prediction $\hat{\zeta}_t\in[0,1]$ estimates the probability that the transition does not terminate because of failure; it subsequently discounts value propagation through imagined trajectories. 
\paragraph{Training process}
First, we pretrain the world model with real macro-transitions collected from the simulator in a supervised fashion. We prefill a replay buffer with real macro transitions generated by an exploratory policy $\pi_{\mathrm{rand}}$ that samples structured latents and guidance variables, and run a fixed number of single-step regression updates. The world model is optimized with the following loss:
\begin{equation}
\begin{aligned}
\mathcal{L}_{\mathrm{WM}}
={}&\frac{1}{d_c}
 \left\lVert\hat{s}_{t+1}^{c}-s_{t+1}^{c}\right\rVert_2^2\\
&+\lambda_f\,\mathrm{BCE}
 \left(\hat{s}_{t+1}^{f},s_{t+1}^{f}\right)
+\lambda_\zeta\,\mathrm{BCE}
 \left(\hat{\zeta}_t,\zeta_t\right).
\end{aligned}
\label{eq:wm_loss}
\end{equation}
Here, the first term is the mean-squared error on the continuous part, the second term is a binary cross-entropy on the discrete task flags, and the last term trains the continuation head. Afterwards, we train the world model and $\pi_H$ concurrently using a shared replay buffer. Each iteration includes three steps: (i) rolling out the current policy in the simulator to collect macro-transitions and adding them to the buffer; (ii) updating the world model on mini-batches sampled from the buffer; and (iii) updating $\pi_H$ on short rollouts imagined by the newly updated world model. Note that the buffer is constantly refreshed with the real experience visited by the current $\pi_H$ at each update. The supplementary material (\emph{Macro-Dynamics World Model}) specifies the complete loss, target construction, replay-buffer procedure, and rollout convention.

\subsection{High-Level Learning in Imagined Dynamics}
\label{sec:hlc}
The goal-conditioned actor $\pi_{H,\varphi}(a_t^H\mid s_t^H,g)$ and critic $V_\xi(s_t^H,g)$ operate at the macro timescale. At each update, real start states are sampled from $\mathcal{B}$ and the actor and WM are unrolled for $Q$ imagined macro steps. The high-level reward $r_t^H=r^H(s_t^H,a_t^H,\hat s_{t+1}^H,g)$ uses the same task-progress definition for real and imagined transitions.
Let $v_t=V_\xi(s_t^H,g)$. We estimate imagined returns using:
\begin{equation}
V_t^{\lambda}=r_t^H+\gamma\hat\zeta_t
\big[(1-\lambda_{\mathrm{ret}})v_{t+1}
+\lambda_{\mathrm{ret}}V_{t+1}^{\lambda}\big],
\quad V_Q^{\lambda}=v_Q,
\label{eq:hlc_lambda}
\end{equation}
where $\lambda_{\mathrm{ret}}$ is the return-trace parameter and $\hat\zeta_t$ downweights returns across predicted termination. Because the high-level action includes sampled discrete decisions, we optimize the actor with a score-function estimator:
\begin{equation}
\mathcal{L}_{\mathrm{actor}}
=
\mathbb{E}_{\hat{\tau}}
\left[
\sum_{t=0}^{Q-1} d_t
\ell_t^{\mathrm{actor}}
\right],
\label{eq:hlc_actor}
\end{equation}

\begin{equation}
\ell_t^{\mathrm{actor}}
=
-\operatorname{sg}(A_t)
\log \pi_{H,\varphi}(a_t^H\mid s_t^H,g)
-\eta_H\mathcal{H}_t,
\label{eq:actor_step}
\end{equation}
where $A_t=(V_t^{\lambda}-v_t)/\max(S,\varepsilon)$ is a return-normalized advantage, $d_t=\prod_{k=0}^{t-1}\gamma\hat\zeta_k$ is the cumulative continuation discount, $S$ is a running return scale, $\varepsilon>0$ prevents division by zero, $\mathcal{H}_t$ is the entropy of the sampled high-level action distribution, and $\operatorname{sg}$ denotes stop-gradient. The critic is trained toward detached $V_t^{\lambda}$ targets with a distributional value loss, while an exponential-moving-average target critic provides a detached distributional regularizer.
\begin{algorithm}[t]
\small
\caption{LUCID training pipeline}
\label{alg:lucid}
\begin{algorithmic}[1]
\REQUIRE Motion data $\mathcal{M}$, simulator $\Psi$,
oracle $\psi$, macro period $K$
\ENSURE Frozen LLC $\pi_L$ and trained HLC $\pi_{H,\varphi}$

\STATE $\displaystyle
\pi_L\leftarrow\arg\max_{\pi}\mathcal{J}_L
(\pi;\mathcal{M},\Psi,\psi)$; freeze $\pi_L$
\STATE Initialize WM $\mathcal{M}_\theta$, actor $\pi_{H,\varphi}$,
critic $V_\xi$, target critic $V_{\bar\xi}$, and replay buffer $\mathcal{B}$

\FOR{$i=1,\ldots,N_{\mathrm{pre}}$}
    \STATE $\tau^{\mathrm{real}}\leftarrow
    \mathsf{Collect}_{K}(\pi_L,\pi_{\mathrm{rand}})$
    \STATE Label visited states with $\psi$ and insert
    $\tau^{\mathrm{real}}$ into $\mathcal{B}$
\ENDFOR

\FOR{$i=1,\ldots,N_{\mathrm{WM}}$}
    \STATE $\theta\leftarrow\theta-\eta_\theta\nabla_\theta
    \mathcal{L}_{\mathrm{WM}}(\theta;\mathsf{Sample}(\mathcal{B}))$
\ENDFOR

\REPEAT
    \STATE $\tau^{\mathrm{real}}\leftarrow
    \mathsf{Collect}_{K}(\pi_L,\pi_{H,\varphi})$
    \STATE Label its visited states with $\psi$ and refresh $\mathcal{B}$

    \STATE $\theta\leftarrow\theta-\eta_\theta\nabla_\theta
    \mathcal{L}_{\mathrm{WM}}(\theta;\mathsf{Sample}(\mathcal{B}))$

    \STATE $\mathcal{D}\sim\mathcal{B}$
    \STATE $\hat{\tau}\leftarrow
    \mathsf{Imagine}_{Q}(\mathcal{D},\pi_{H,\varphi},\mathcal{M}_\theta)$

    \STATE $\xi\leftarrow\xi-\eta_\xi\nabla_\xi
    \mathcal{L}_{\mathrm{critic}}(\xi;\hat{\tau})$

    \STATE $\varphi\leftarrow\varphi-\eta_\varphi\nabla_\varphi
    \left[
      \mathcal{L}_{\mathrm{actor}}(\varphi;\hat{\tau})
      +\beta_{\mathrm{BC},k}\mathcal{L}_{\mathrm{BC}}(\varphi;\mathcal{D},\psi)
    \right]$

    \STATE $\bar{\xi}\leftarrow
    (1-\alpha)\bar{\xi}+\alpha\xi$;\quad
    $\beta_{\mathrm{BC},k+1}\leftarrow\mathsf{Decay}(\beta_{\mathrm{BC},k})$
\UNTIL{the training budget is exhausted}

\RETURN $\pi_L,\pi_{H,\varphi}$
\end{algorithmic}
\end{algorithm}
To stabilize early skill discovery, we warm-start the categorical skill-selection component from a geometry-based oracle. Its behavior-cloning coefficient is annealed to zero, so the oracle initializes rather than fixes the final controller. 

Algorithm~\ref{alg:lucid} summarizes the full optimization procedure for LUCID. Here, $\mathcal{J}_L$ is the LLC objective in Eq.~\eqref{eq:llc_objective}, $\mathcal{M}_\theta$ is the macro-dynamics world model, and $\theta$, $\varphi$, and $\xi$ denote the WM, actor, and critic parameters, respectively. The oracle-labeled buffer is used only while $\beta_{\mathrm{BC},k}>0$.
The exact joint action distribution, critic loss, oracle objective, and actor--WM update schedule are provided in the supplementary material (\emph{High-Level Controller Training}).

\section{Experiments}
We evaluate LUCID on long-horizon multi-object rearrangement, focusing on three questions: whether it improves sequential task completion over representative baselines, whether it remains effective on held-out layouts and longer task chains, and which components account for its performance. We first define the shared evaluation protocol, then report comparison results for diverse settings, followed by world-model and latent-interface ablations.
\subsection{Experimental Setup}
\subsubsection{Dataset} 
We construct multi-object rearrangement tasks from HITR~\cite{xu2024humanvla}. Movable objects are initialized on the ground in diverse physically feasible configurations across warehouse, kitchen, bedroom, and living-room layouts. We compose single-object instances into sequential episodes in which each object must be placed before the next subtask becomes active. The ID split contains 62 training tasks, whereas the OOD split contains 20 tasks. The standard benchmark contains two-object chains; a separate extension evaluates chains of up to five objects. LLC training additionally uses reference motions from OMOMO~\cite{li2023object} and SAMP~\cite{hassan2021stochastic}.
\subsubsection{Metrics}
We report three metrics. \textbf{Success Rate} SR$_k$ is the percentage of episodes in which the first $k$ objects are placed within $0.2\,\mathrm{m}$ of their targets. \textbf{Average Placement Error} (APE) is the final object-to-goal distance averaged over all task objects. \textbf{Time} is the mean simulated completion time over successful episodes. Values are mean $\pm$ sample standard deviation over three evaluation seeds using the same task split and success criterion for every method.
\subsubsection{Baselines} 
We compare against the following baselines using task-specific deployment adapters. \textbf{InterMimic}~\cite{wang2025skillmimic} is a motion-tracking policy; for each subtask, we construct a kinematic human--object reference by warping motion segments to the required waypoints and object poses. \textbf{TokenHSI}~\cite{pan2025tokenhsi} uses a scripted finite-state adapter to switch between locomotion and object-carrying skills. \textbf{HumanVLA}~\cite{xu2024humanvla} is an end-to-end single-object rearrangement policy, which we extend by activating subtasks sequentially. All methods use the same ID/OOD task lists, simulator horizon, placement threshold, and evaluation seeds. Because these baselines do not model autonomous inter-subtask transitions, their adapters reset only the humanoid to a favorable pose at each handoff. Consequently, SR and APE compare LUCID against favorably initialized baseline deployments, while completion times are descriptive and not directly comparable.
\subsection{Implementation Details}
We use Isaac Gym~\cite{makoviychuk2021isaac} and a humanoid with 15 rigid bodies and 28 PD-controlled joints~\cite{hassan2023synthesizing}. The LLC follows ASE but replaces its learned skill encoder and diversity objective with the structured latent interface. We train it with PPO~\cite{schulman2017proximal} and a staged curriculum that progresses from robust single-object interaction to two-object chaining. The HLC uses a DreamerV3-style actor--critic, and the deterministic world model is a plain MLP over the compact task state. The simulator runs at $60\,\mathrm{Hz}$, the LLC acts at $30\,\mathrm{Hz}$, and the HLC selects one macro action every 20 LLC steps. The HLC uses a 12-step imagination horizon, sequence length 32, and batch size 64. We train the LLC on two NVIDIA RTX 4090 GPUs and train the HLC and world model jointly with 8192 parallel environments. Further architecture, optimization, and evaluation details are provided in the supplementary material.
\begin{table*}[t]
\centering
\small
\setlength{\tabcolsep}{3.5pt}
\begin{tabular}{lcccccccc}
\toprule
& \multicolumn{4}{c}{\textbf{In-Distribution}} &
\multicolumn{4}{c}{\textbf{Out-of-Distribution}} \\
\cmidrule(lr){2-5}\cmidrule(lr){6-9}
\raisebox{1.5ex}[0pt][0pt]{\textbf{Method}} & SR$_1$ ($\%$) $\uparrow$ & SR$_2$ ($\%$) $\uparrow$ & APE (m) $\downarrow$ & Time (s) $\downarrow$ &
SR$_1$ ($\%$) $\uparrow$ & SR$_2$ ($\%$) $\uparrow$ & APE (m) $\downarrow$ & Time (s) $\downarrow$ \\
\midrule
HumanVLA & $79.4\pm0.7$ & $39.8\pm0.8$ & $0.83\pm0.02$ & $6.1\pm0.1$
& $77.8\pm5.6$ & $37.0\pm0.8$ & $0.77\pm0.05$ & $6.2\pm0.4$ \\
InterMimic & $23.7\pm4.6$ & $4.7\pm1.2$ & $2.21\pm0.04$ & $10.4\pm1.3$
& $19.4\pm7.3$ & $3.7\pm3.2$ & $2.11\pm0.23$ & $11.9\pm1.9$ \\
TokenHSI & $69.4\pm1.1$ & $28.7\pm1.9$ & $0.96\pm0.15$ & $7.6\pm0.3$
& $66.1\pm3.7$ & $26.4\pm3.5$ & $1.17\pm0.09$ & $8.2\pm0.2$ \\
\textbf{LUCID} & $89.2\pm0.7$ & $73.4\pm3.0$ & $0.32\pm0.02$ & $15.4\pm0.9$
& $83.9\pm0.8$ & $68.4\pm3.8$ & $0.52\pm0.05$ & $18.3\pm2.4$ \\
\bottomrule
\end{tabular}
\caption{Comparison on the shared ID and OOD task splits. SR$_k$ denotes completion through subtask $k$.}
\label{tab:general_comparison}
\end{table*}
\begin{figure*}[t]
    \centering
    \includegraphics[width=0.95\textwidth]
    {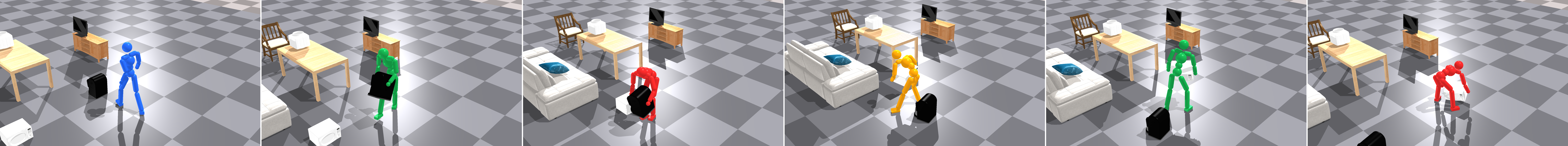}
    \vspace{1mm}
    \includegraphics[width=0.95\textwidth]
     {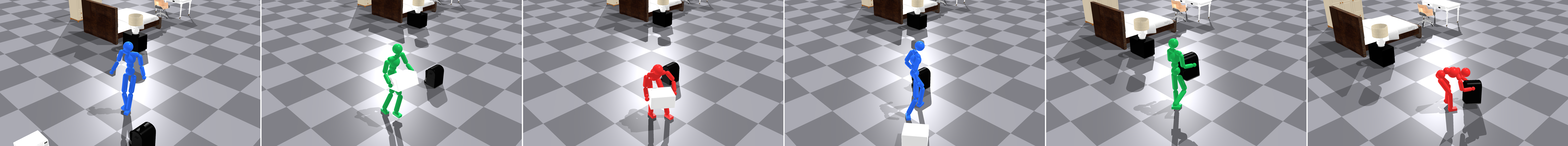}
    \vspace{1mm}
    \includegraphics[width=0.95\textwidth]
     {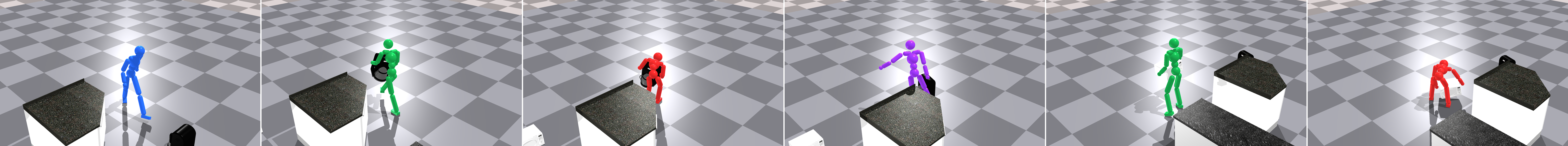}
    \vspace{1mm}
    \includegraphics[width=0.95\textwidth]
     {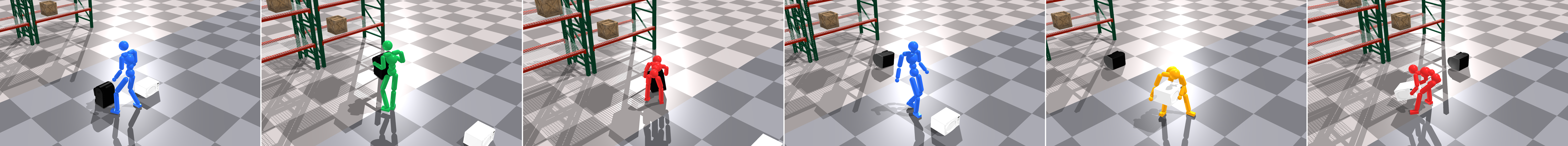}
\caption{Qualitative filmstrips of LUCID in various task layouts (top to bottom: living room, bedroom, kitchen, warehouse). Colors indicate semantic latent decisions: locomotion (blue), transport (green), release (red), fetch (yellow), and transit (purple). The frames illustrate continuous navigation, object interaction, and autonomous handoff without resetting the humanoid.}
\label{fig:qualitative_filmstrips}
\end{figure*}
\begin{figure}[t]
\centering
\includegraphics[width=0.43\textwidth]{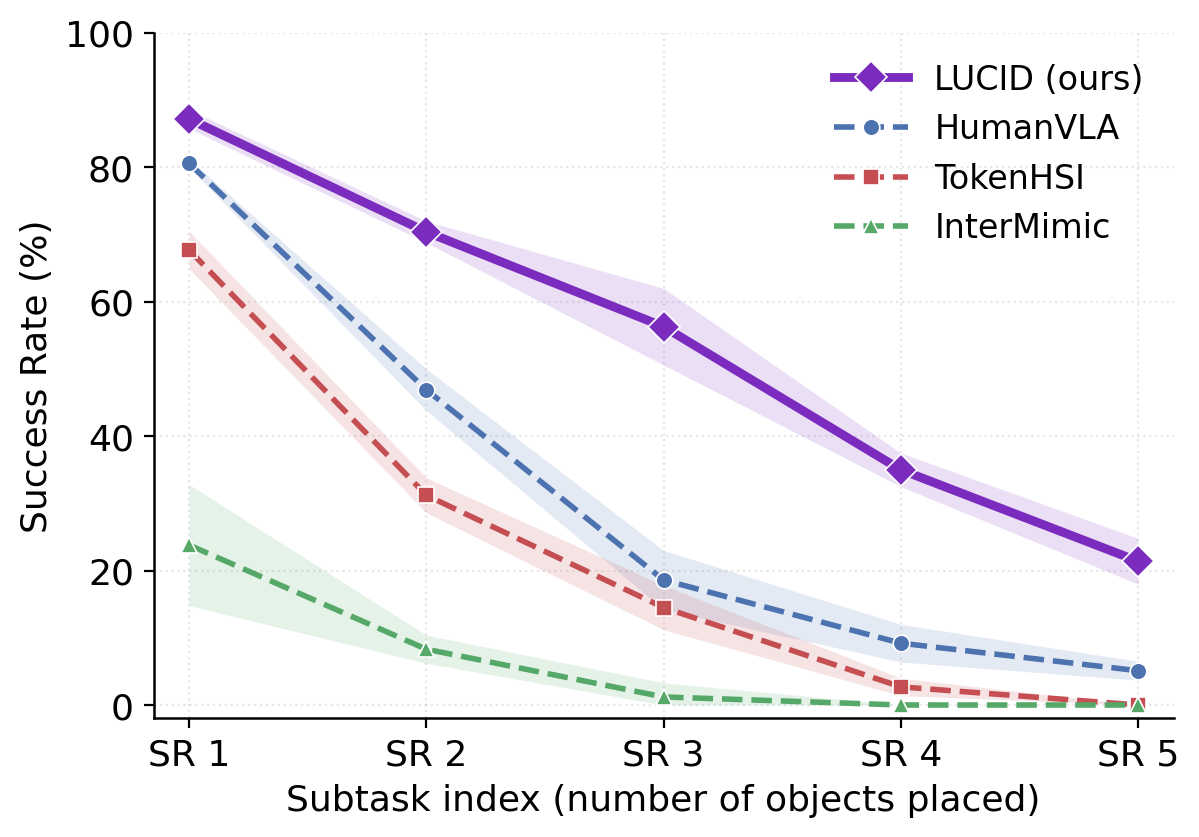} 
\caption{Success rates against the number of objects.}
\label{fig:extension_comparison}
\end{figure}
\begin{figure}[t]
\centering
\includegraphics[width=0.45\textwidth]{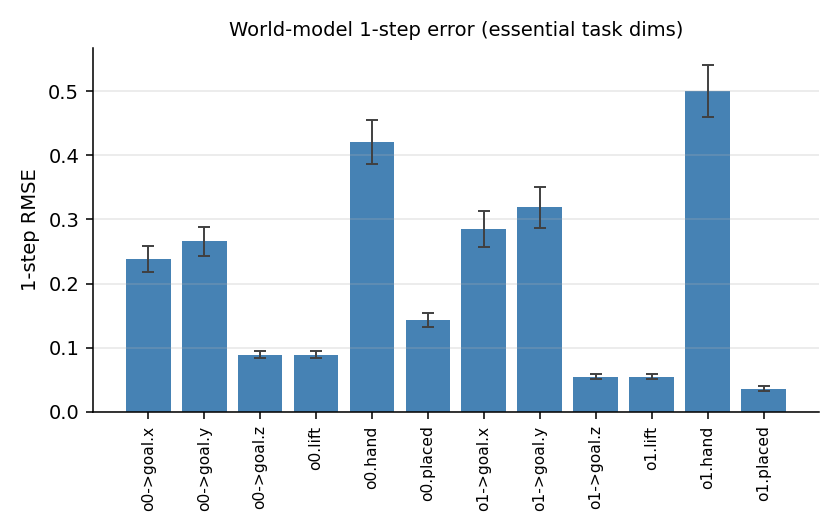} 
\caption{World-model one-step RMSE on task-relevant compact-state dimensions. Bars average per-environment RMSE and whiskers denote the standard error across evaluation environments.}
\label{fig:wm_per_dim_rmse}
\end{figure}
\begin{figure}[t]
\centering
\includegraphics[width=0.45\textwidth]{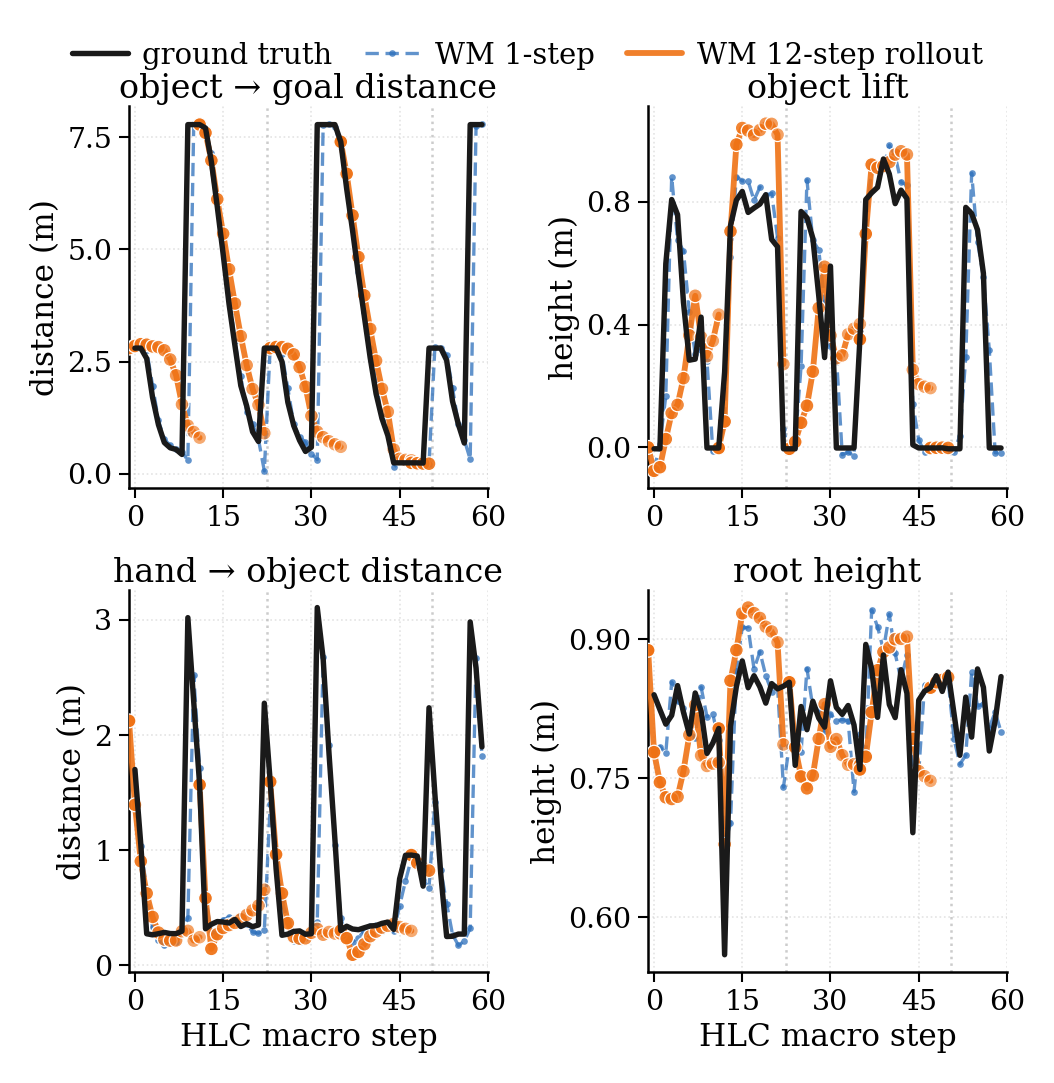} 
\caption{Multi-step world-model predictions. Black curves show simulator states, dashed blue curves show one-step predictions, and orange curves show imagined rollouts.}
\label{fig:wm_traj_pred}
\end{figure}
\begin{figure}[t]
\centering
\includegraphics[width=0.48\textwidth]{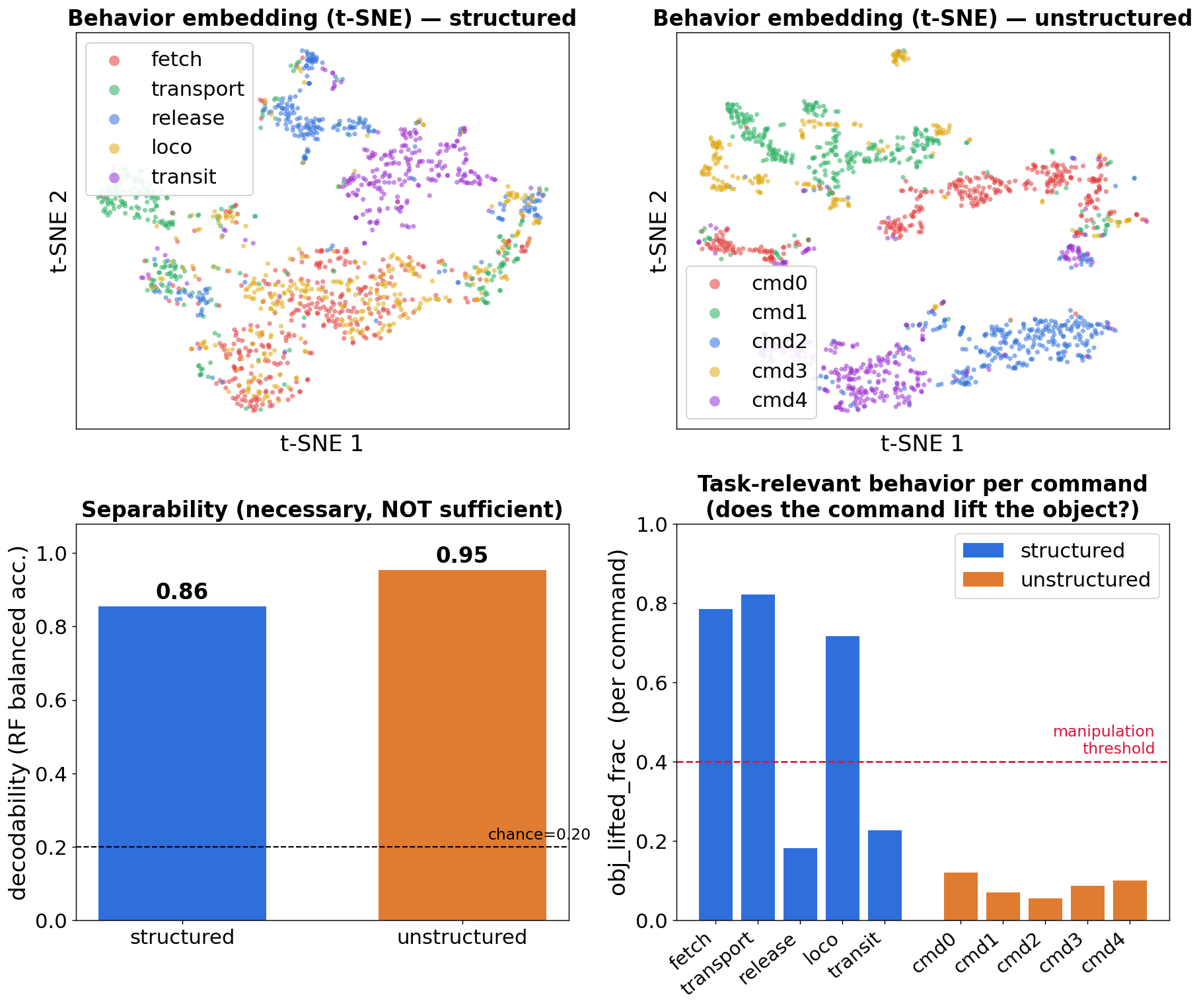} 
\caption{Behavior controllability of the structured and unstructured latent interface. Top: separately fitted t-SNE visualizations of identical kinematic descriptors. Bottom left: five-fold random-forest balanced accuracy for command decoding. Bottom right: mean fraction of time each command lifts the object, with $0.4$ used as the sustained threshold.}
\label{fig:structured_latent}
\end{figure}
\begin{table}[t]
\centering
\small
\setlength{\tabcolsep}{2.6pt}
\renewcommand{\arraystretch}{1.15}
\begin{tabular}{@{}lcccc@{}}
\hline
& \multicolumn{2}{c}{\textbf{ID}} & \multicolumn{2}{c}{\textbf{OOD}} \\[-1pt]
\cline{2-3}
\cline{4-5}
\raisebox{0.85ex}[0pt][0pt]{\textbf{Method}}
& SR1 ($\%$) $\uparrow$ & SR2 ($\%$) $\uparrow$ & SR1 ($\%$) $\uparrow$ & SR2 ($\%$) $\uparrow$ \\
\noalign{\vskip 2.5pt}
\hline
\noalign{\vskip 3pt}
MF-HLC-sparse
& $84.8\pm0.7$ & $59.3\pm1.3$ & $82.6\pm0.8$ & $42.8\pm1.1$ \\[1.5pt]
MF-HLC-dense 
& $87.8\pm0.6$ & $61.9\pm1.1$ & $65.0\pm2.0$ & $33.1\pm1.2$ \\
\noalign{\vskip 3pt}
\hline
\noalign{\vskip 3pt}
LUCID (\emph{sparse})
& $89.5\pm1.0$ & $73.5\pm3.0$ & $81.9\pm0.9$ & $63.1\pm2.7$ \\[1.5pt]
LUCID-dense
& $90.8\pm0.7$ & $74.0\pm1.0$ & $70.1\pm0.9$ & $50.2\pm3.2$ \\
\noalign{\vskip 3pt}
\hline
\end{tabular}
\caption{WM ablation under sparse and dense rewards.}
\label{tab:wm_mf_comparison}
\end{table}
\begin{table}[t]
\centering
\small
\setlength{\tabcolsep}{2.6pt}
\renewcommand{\arraystretch}{1.15}
\begin{tabular}{@{}lcccc@{}}
\hline
& \multicolumn{2}{c}{\textbf{ID}} & \multicolumn{2}{c}{\textbf{OOD}} \\[-1pt]
\cline{2-3}
\cline{4-5}
\raisebox{0.85ex}[0pt][0pt]{\textbf{Method}}
& SR1 ($\%$) $\uparrow$ & SR2 ($\%$) $\uparrow$ & SR1 ($\%$) $\uparrow$ & SR2 ($\%$) $\uparrow$ \\
\noalign{\vskip 2.5pt}
\hline
\noalign{\vskip 3pt}
LUCID (\emph{struct})
& $87.5\pm1.5$ & $74.2\pm2.6$ & $82.0\pm1.5$ & $66.4\pm3.6$ \\[1.5pt]
LUCID-unstruct
& $4.8\pm2.6$ & $0.0\pm0.0$ & $2.7\pm1.9$ & $0.0\pm0.0$ \\
\noalign{\vskip 3pt}
\hline
\end{tabular}
\caption{Structured-interface ablation. Each HLC is trained and evaluated with its corresponding frozen LLC interface under the same ID/OOD protocol.}
\label{tab:structured_latent}
\end{table}
\subsection{Comparison Experiments}
\subsubsection{Quantitative evaluation}
Table~\ref{tab:general_comparison} compares all methods on the shared ID and OOD protocols. LUCID obtains the highest mean SR$_1$, SR$_2$, and the lowest APE on both splits despite receiving no handoff reset. On ID tasks, its SR$_2$ exceeds the strongest baseline by $33.6$ percentage points ($73.4\%$ versus $39.8\%$); on OOD tasks, the margin is $31.4$ points ($68.4\%$ versus $37.0\%$). The smaller SR$_1$--SR$_2$ gap for LUCID ($15.8$ points ID and $15.5$ points OOD) indicates more reliable handoffs than the baselines. From ID to OOD, LUCID decreases by $5.3$ points in SR$_1$ and $5.0$ points in SR$_2$, while retaining the best mean placement accuracy. InterMimic performs worst overall, consistent with the difficulty of constructing precise warped references for diverse object configurations. Baseline completion times are shorter because their handoff adapters teleport the humanoid; we therefore treat Time as descriptive rather than a direct efficiency comparison.

\subsubsection{Qualitative performance}
 Figure~\ref{fig:qualitative_filmstrips} shows LUCID completing multi-object rearrangement chains in diverse layouts. We can see that LUCID exhibits clean phase structures and well-separated transitions across all scenes, with HLC producing interpretable skills that are faithfully executed by the frozen LLC. The same latent repertoire generalizes across layouts without per-scene tuning, adapting to environmental obstacles while completing the task chain fully autonomously. These are consistent with the SR2 and placement-accuracy gains reported in Table~\ref{tab:general_comparison}. The supplementary material provides the corresponding evaluation protocol and implementation details.

\subsubsection{Extended multi-objects tasks}
To assess long-horizon robustness beyond the two-object training horizon, we evaluate chains of up to five objects and report prefix success SR$_k$ in Figure~\ref{fig:extension_comparison}. InterMimic and TokenHSI approach zero by SR$_4$, and HumanVLA retains $5\%$ success at SR$_5$. This rapid decay reflects their reliance on scripted planners, which accumulate handoff errors without recovery. However, LUCID degrades more gracefully, achieving around $56\%$ at SR3 and still $21\%$ at SR5. We attribute this to the joint evolution of the world model and high-level policy during training, which together yields a more reliable long-horizon planner. 

\subsection{Ablation Studies}
We investigate the key components of LUCID by addressing the following questions: 
1) \emph{Does the world model improve HLC learning through imagined rollouts?}
2) \emph{Does the structured latent interface improve behavior controllability?}
\subsubsection{On world model effectiveness}
We first evaluate predictive accuracy independently of downstream task success. Figure~\ref{fig:wm_per_dim_rmse} reports one-step errors for task-critical dimensions, including object--goal displacement, lift height, hand--object distance, and placement progress. Figure~\ref{fig:wm_traj_pred} then visualizes a representative 12-step open-loop rollout. The supplementary material specifies the action sequence, sampling protocol, and aggregate horizon-wise errors used alongside this qualitative trajectory.
Table~\ref{tab:wm_mf_comparison} compares HLCs trained with world-model (WM) imagination against model-free (MF) learning under both sparse and dense reward settings. The WM variants achieve higher mean SR$_2$ in all four split--reward comparisons: $73.5$ versus $59.3$ and $74.0$ versus $61.9$ on ID, and $63.1$ versus $42.8$ and $50.2$ versus $33.1$ on OOD. This consistent advantage supports the use of imagined macro-transitions for learning long-horizon task progression. Dense shaping yields only a $0.5$-point ID SR$_2$ gain for LUCID but reduces OOD SR$_2$ by $12.9$ points, suggesting weaker transfer than the sparse setting.

\subsubsection{On structured latent interface}
Figure~\ref{fig:structured_latent} separates command decodability from task relevance. We drive each frozen LLC with five commands in the same two-object environment, retain states in which the object is initially reachable, and describe the resulting motions with identical kinematic features. The independently fitted t-SNE plots and five-fold random-forest probe show that both interfaces produce command-decodable behaviors (balanced accuracy $0.86$ and $0.95$). Decodability alone is therefore insufficient. The manipulation panel instead shows that three structured commands exceed the sustained-lift threshold, whereas none of the unstructured commands does. Table~\ref{tab:structured_latent} further shows that the structured interface reaches $74.2\%$ ID and $66.4\%$ OOD SR$_2$, while the unstructured interface obtains $0.0\%$ SR$_2$ on both splits. These results indicate that the relevant property is not merely diverse behavior, but a command set aligned with the interaction stages required by the downstream task.


\section{Conclusion}
\label{sec:conclusion}
In this paper, we introduced LUCID, a hierarchical framework for multi-object rearrangement with a simulated humanoid. It coordinates a structured low-level controller using a high-level policy trained through imagined skill-level transitions, improving long-horizon completion. Results show that task-aligned temporal abstraction is central: a structured skill interface makes high-level decisions meaningful, while macro-level prediction supports long-horizon reasoning without reproducing joint-level dynamics. Current limitations include privileged state inputs and limited task, object, and embodiment diversity; future work will address visual control, broader tasks, and real-world transfer.

\bibliography{aaai2027}
\end{document}